\documentclass[a4paper,article,10pt]{memoir}
\usepackage[english]{babel}
\usepackage{CCLAuthor}
\usepackage[reqno]{amsmath}% AMS-LaTeX-first 
\usepackage{amssymb}
\usepackage{amsthm}  
\theoremstyle{plain}

\theoremstyle{definition}

\numberwithin{theorem}{chapter}% AMS-LaTeX-last
\usepackage[round,authoryear]{natbib} % bibliography
\usepackage[]{graphicx}               % graphics
\begin{document}
%
%----------------------------------------------
%
%    declare the TITLE of your contribution
%
\title{Design of a robot for the automatic charging of an electric car}
%    
%    AUTHOR(s) and AFFILIATION(s)
%
\author{%
    Damien Chablat\textsuperscript{\dag\ddag}
       and 
     Riccardo Mattacchione\textsuperscript{\ddag}
        and
     Erika Ottaviano\textsuperscript{\ddag}
    \\ \smallskip\small% some space
    % begin affiliations
    \textsuperscript{\dag} % second author
    DICeM, University of Cassino
G. Di Biasio 43, 03043 Cassino (Fr), Italy
    \\ 
    \textsuperscript{\ddag} % a common affiliation
    LS2N, UMR CNRS 6004, 1 rue de la Noë,
    44321 Nantes, France
    }% end affiliations
%   
%   typeset the title
%   
    \maketitle
%
%----------------------------------------------
%
%   ABSTRACT: very important for database indexing!
%   ap140610
%
\noindent \begin{abstract}
In this paper, a robot with parallel architecture is proposed for charging an electric vehicle having the charging socket on its front side. Kinematic models are developed to design the robot for a given workspace that corresponds to the car's plug placements. A demonstrator composed by commercial components is shown.  
\end{abstract}
% end abstract
%%%%%%%%%%%%%%%%%%%%%%%%%%%%%%%%%%%%%
\CCLsection{Introduction}
%%%%%%%%%%%%%%%%%%%%%%%%%%%%%%%%%%%%%
Nowadays, the automotive industry develops electric vehicles with associated tools for a broad range of customers and users. To charge the batteries, most users still have to manually connect the car to a charging station or home outlet. In addition, to enhance the range of electric cars, the batteries are equipped with a larger capacity and require a high charging intensity, which must be performed by large cables. Approximately one meter of cable consisting of four wires with a section of $25mm^2$ each has a mass of approximately 0.9~Kg. Therefore, some pioneering studies have been developed to explore the possibility of using a robot that could relieve the user from the task of log in and out the plug. However, among the other possible industrial applications in the automotive industry, only a few robots are designed exclusively for recharging electric cars. In August 2015, the car manufacturer Tesla presented a snake robot for charging Tesla Model S cars. This robot has the morphology of a snake and can detect the vehicle's charging port and connect to it autonomously. The charging cover opens automatically when the parked vehicle is ready to be charged. Once the connection between the robot and the vehicle is established, the charging process can be started. The idea behind this concept is that the driver does not need to get out of the car to charge the vehicle  (\cite{walzel2016automated}). 

In July 2017, Wolkswagen company unveiled discreetly the prototype of the future Gen.E. A robotic articulated arm equipped with a combo plug is installed on the side of the vehicle. The robot is designed for underground and multi-store garages for autonomous charging of the vehicle. Except the snake robot, whose architecture is unique, the mechanical architecture of the robot for charging the Gen.E is the KUKA serial architecture. The end-effector is adapted for the insertion of the power outlet into the electric vehicle. The DC quick charging process starts with communication between the vehicle and the electric filling station. The vehicle transmits its data to the charging station, which gives back the target position for the automated parking (\cite{walzel2016automated}). The charging socket of the vehicle has to be in a target area of 200 by 200 mm. Afterwards, a camera on the robot detects the exact position of the charging socket, with an accuracy of mm. After, the gripper picks up the DC-Connector and connects with the charging socket of the vehicle. After having linked the DC-Connector, the charging process starts. Once the battery is fully charged, the robot automatically unplugs the DC-Connector. 

Another automatic recharge system is the NRG-X, it can be adapted to any EV or PHEV and is capable of fast charging. Furthermore, it has a tolerance for inaccurate parking positions. The NRG-X system is based on a combination of conductive and inductive charging beneath the vehicle, thus an adapter for the vehicle is necessary (\cite{miseikis20173d}). It has to be noted that although  industrial robots can very well adapt to several tasks, their design and production remain very expensive for the application. 

For public places, such as car parks, shared solutions have been proposed (\cite{walzel2016automated}). However, this solution is expensive in terms of infrastructure and cannot be used for individual houses. The price of an industrial robot is about 20-40\% of the price of a car. 
Research projects in this context aim to develop specific robots by reducing the number of actuators (\cite{yuan2020concept}) or overcoming the effects of gravity on actuators (\cite{gao2016robotically}) and (\cite{KetfiCherif2018}), these are the main objectives on which this article is based.
This is the main reason why the French company Renault decided in 2018-2019 to develop a less expensive charging robot for its upcoming electric cars (\cite{Kamga:2017}). The specifications for the robot indicate that it has to be mounted on the garage wall and will recharge the battery by facing the car. The robot must be small, precise, and stiff. It must allow the plug to be inserted in safely and properly way. The main requirements are: (a) Insertion and withdrawal forces: 70 N, ±10 N, (b) Depth of insertion: 40 mm, ±20 mm, (c) Distance between plug and socket: 20 mm, +10 mm, (d) Workspace: 200 ×200 ×20 $mm^3$, (e) Degree of freedom: 3, (f) Location accuracy: 0.5 mm, ±0.1 mm, (g) Length of electric cables: 1500 mm, +500 mm, (h) Weigh of electric cables: 1.4 kg,+0.45 kg, (i) Plug diameter: 65 mm, ±5 mm, (j) Total weigh of the robot: 10 kg, +5 kg, and (k) Occupied space 1500 ×200 $mm^2$.
%%%%%%%%%%%%%%%%%%%%%%%%%%%%%%%%%%%%%%
\CCLsection{Design requirements of a robot for the automatic charging of an electric car}
%%%%%%%%%%%%%%%%%%%%%%%%%%%%%%%%%%%%%%
Our work focuses on the Renault Zoe, for this car, the plug is placed on the front and is slightly inclined. The mobility required for the robot could be at least three translations if the car is on a horizontal floor, plus a translation along the insertion axis for the electric plug. By using the flexibility of the insertion tool, it is possible to avoid using a wrist or spatial mechanisms, (\cite{FiglioliniAsme2012}), (\cite{FiglioliniRCC2016}).

\begin{figure}[htbp]
    \centering
    \includegraphics[width = 0.55\textwidth]{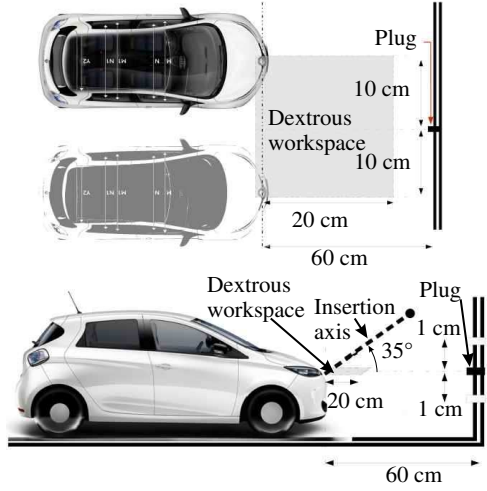}
    \caption{Workspace of robot and the Zoe car}
    \label{fig:robot_car}
\end{figure}

To reduce the cost of the robot, it is possible to mix the vertical movement with the insertion direction.
Figure~\ref{fig:robot_car} depicts the car and the dexterous workspace of the robot.
Several robot architectures were analyzed  before settling on a parallel \underline{P}-Pi-R-\underline{P} type robot for planar motion (\cite{Rea-IR2013}) put in series with a linear actuator for insertion movement where \underline{P} stands for prismatic actuated joints, Pi for a four-bar mechanism similar to an RR mechanism but with opposite bars parallel two by two, and R for revolute joint. 

The main advantages of this architecture are, simplified design (\cite{wenger2001comparative,OttavianoRobotica2013}), and the forces applied to the actuators decrease as the end-effector moves away from the base. This will be the case when inserting the plug.
%%%%%%%%%%%%%%%%%%%%%%%%%%%%%%%%%%%%%%
\CCLsection{Mechanism understudy}
%%%%%%%%%%%%%%%%%%%%%%%%%%%%%%%%%%%%%%
%The kinematics of the robot can be written separately, the $\alpha$ and $\beta$ mechanism, before assembly in a single one the $\gamma$ mechanism. 
The kinematics of the robot can be written in three stages, (i) the $alpha$ mechanism which allows translations of the end-effector in one plane, (ii) the $beta$ mechanism which does the insertion of the plug, and (iii) the $gamma$ mechanism which is the assembly of the first two. 
%%%%%%%%%%%%%%%%%%%%%%%%%%
\CCLsubsection{The $\alpha$ mechanism: P-Pi-R-P planar parallel robot}
%%%%%%%%%%%%%%%%%%%%%%%%%%
The first mechanism aims to achieve a planar displacement. To choose this architecture, the constraint of storage on a wall and the reduction of on-board weight lead to the choice of prismatic joints.

The design parameters of the $\alpha$ mechanism are the following:
\begin{itemize}
    \item	$L$: the length of the two arms; 
    \item	$L_2$: the length of the robot’s base;
    \item	$L_3$: the distance between $B$ and $C$ 
    \item	$\rho_1$ and $\rho_2$: the length of the left and right linear actuator;
    \item	$[x, y]$: the coordinate of point $V$ located in the middle of ($B$, $C$);
    \item   $[0, e_{py}]$: the offset from $V$ to $F$; 
    \item	$[e_{1x}+\rho_1, e_{1y}]$: the coordinate of $A$;
    \item	$[L_2 - \rho_2 - e_{2x},e_{2y}]$: the coordinate of $E$;
\end{itemize}
\begin{figure}[htbp]
    \centering
    \includegraphics[width = 0.9\textwidth]{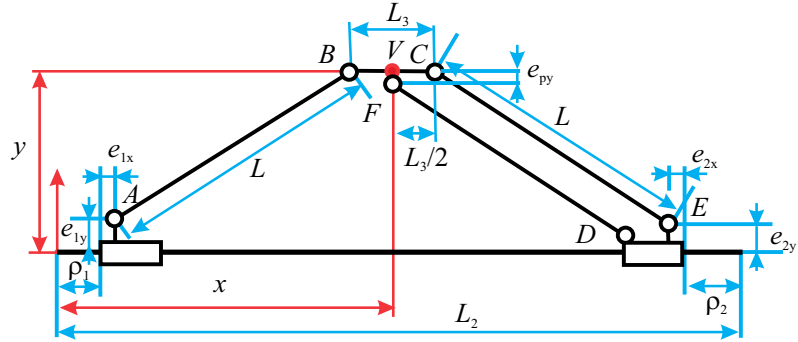}
    \caption{Kinematic scheme of the $\alpha$ mechanism}
    \label{fig:alpha_mech}
\end{figure}

According to the kinematic scheme in Figure~\ref{fig:alpha_mech}, the constraint equations that can be obtained are the following
\begin{eqnarray}
(x-e_{1x}-\rho_1-L_3/2)^2+(y -e_{1y})^2&=&L^2 \label{system1}\\
(L_2-x-L_3/2-e_{2x}-\rho_2)^2+(y-e_{2y})^2&=&L^2 \label{system2}
\end{eqnarray}

with Eqs.~\ref{system1}--\ref{system2}, the inverse kinematic equations can be obtained as
\begin{eqnarray}
\rho_1&=& x-e_{1x}-L_3/2-\sqrt{2L^2-(y-e_{1y})^2} \\
\rho_2&=& L_2-x-L_3/2-e_{2x}-\sqrt{2L^2-(y-e_{1y})^2}			
\end{eqnarray}
Solving Eqs.~\ref{system1}--\ref{system2} for the Cartesian coordinates $x$ and $y$  allows the evaluation of two assembly modes. However, only one will be accessible, thanks to the joint limits (\cite{wenger2001comparative}). 

To avoid the singularity of the Pi joint in the folded position, the offset $e_{py}$ is defined. The inclined placement of the parallelogram allows sufficient stiffness to be maintained when the robot is close to its base in the folded position. Indeed, the stiffness of a parallelogram is proportional to the area formed by the four bars (\cite{majou2002design}). The placement of the bar $(DF)$ does no influence on the kinematics, but it does influence the stiffness of the robot.

\begin{figure}[htbp]
    \centering
    \includegraphics[width = 0.9\textwidth]{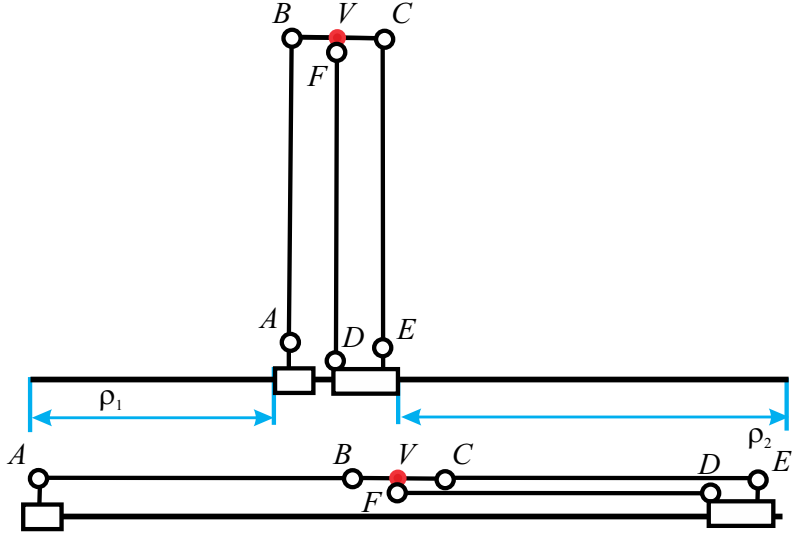}
    \caption{Parallel and serial singular configuration of the $\alpha$ mechanism}
    \label{fig:alpha_mech_sing}
\end{figure}

In Figure~\ref{fig:alpha_mech_sing}, two singular postures are depicted to define the boundaries of the workspace and to separate into two aspects by the parallel singularity. The IIM-type singularity of the four-bar mechanism defined in (\cite{zlatanov1993singularity}) is avoided thanks to the offset $e_{py}$.

\begin{figure}[htbp]
        \centering
        \includegraphics[width = 0.75\textwidth]{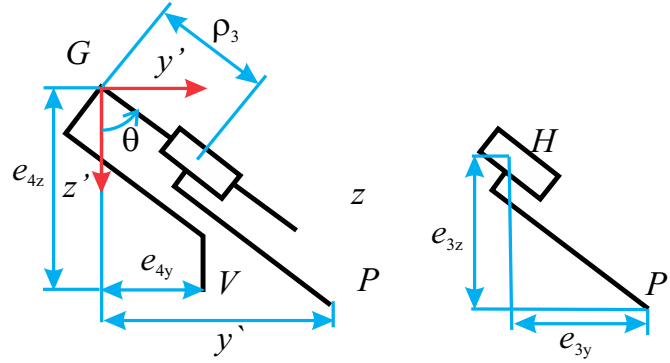}
        \caption{Kinematic scheme of the $\beta$ mechanism.}
	    \label{fig:beta_mechanism}
    \end{figure}

\begin{figure}[t]
  \begin{minipage}[b]{0.50\linewidth}
  \center
    \includegraphics[width = 0.95\textwidth]{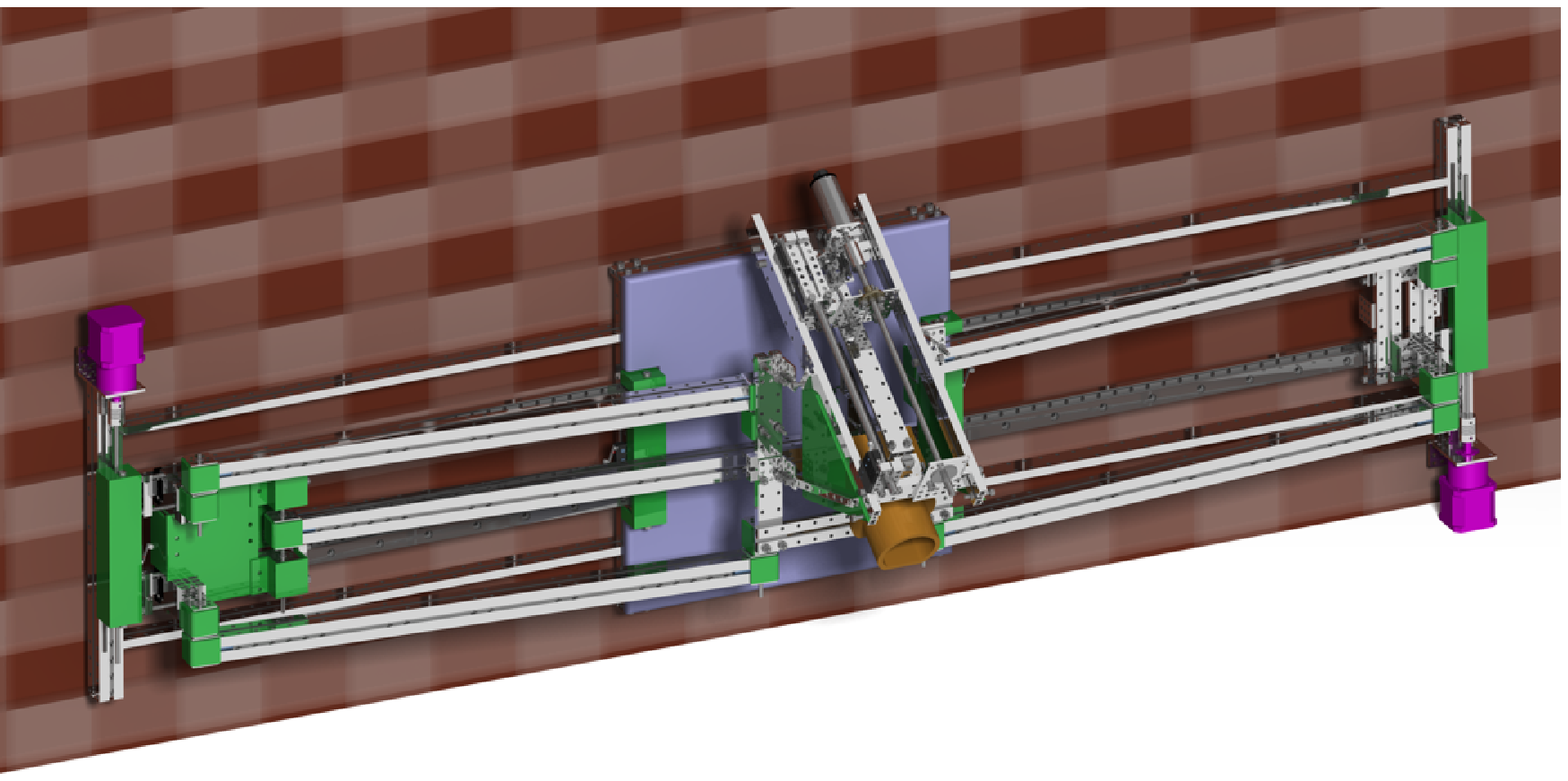}
\caption{CAD mode of robot}
    \label{fig:robot_CAD}
  \end{minipage} 
  \begin{minipage}[b]{0.5\linewidth}
    \includegraphics[width = 0.95\textwidth]{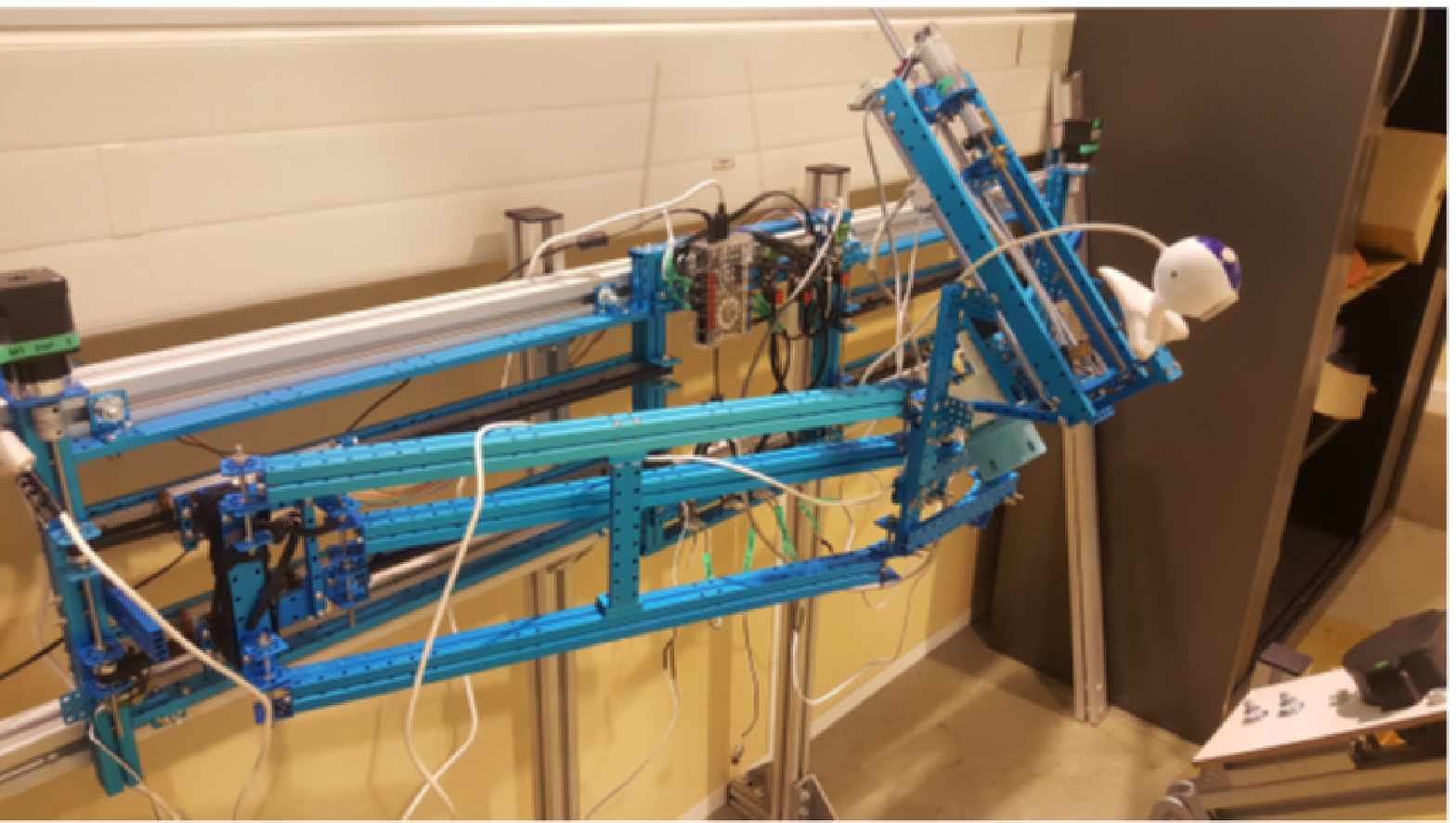}
    \caption{Prototype of the robot}
    \label{fig:robot}
  \center
 \end{minipage}
\end{figure}

%%%%%%%%%%%%%%%%%%%%%%%%%%%%%%%%%%%%%%
\CCLsubsection{The $\beta$ mechanism: serial robot}
%%%%%%%%%%%%%%%%%%%%%%%%%%%%%%%%%%%%%%
For this mechanism, the constraint equations define $P$ with respect to the Cartesian frame  $z'$-$y'$. According to Figure~\ref{fig:beta_mechanism}, it is possible to note that:
    \begin{itemize}
\item 	$\rho_3$: is the length of the linear actuator that defines the coordinate of $H$ with respect to $G$; 
\item 	$[e_{3y},e_{3z}]$: is the coordinate of $H$ with respect to $P$;
\item 	$[e_{4y}, e_{4z}$: is the coordinate of $V$ in the reference frame $z'$-$y'$.  
\item 	$[x', y'$: is the coordinate of $P$ in the reference frame $z'$-$y'$.  
    \end{itemize}
According to the kinematic scheme in Figure~\ref{fig:beta_mechanism}, the constraint equations that can be obtained are the following
\begin{eqnarray}    
    z'&=& e_{3z} + \rho_3 \cos(\theta)\\
    y'&=& e_{3y} + \rho_3 \sin(\theta)
\end{eqnarray}
and, its inverse kinematic as
\begin{equation}
    \rho_3 =\sqrt{(z'-e_{3z})^2+(y'-e_{3y})^2}
\end{equation}
The offset actuator reduces the volume of the robot when it is folded\footnote{See movie: https://youtu.be/lCRXJ5tpTMM}.
%%%%%%%%%%%%%%%%%%%%%%%%%%%%%%%%%%%%%%
\CCLsubsection{The $\gamma$ mechanism}
%%%%%%%%%%%%%%%%%%%%%%%%%%%%%%%%%%%%%%
The merging of the mechanisms described in earlier subsections gives a 3 d.o.f. robot, the $\gamma$ mechanism. For verifying the operation of the robot prototype, an Arduino micro-controller can be used.  Figure~\ref{fig:robot_CAD} shows a CAD model in CATIA V5 with components from \cite{makeblock}. 
%The purpose of this subsection is to provide an analytical calculation of the three equations for the position analysis of  $P(X, Y, Z)$. First, the next assumptions must be considered, according to Figure~\ref{fig:robot}.
%\begin{eqnarray}
%	X &=& x \\
%	Y &=& y + y'-e_{4y} \\
%	Z &=& z = z' -e_{4z}
%\end{eqnarray}

%    
%From Equation XX, direct kinematic equations can be obtained as  
%\begin{eqnarray}
%X&=&\frac{-Q \pm \sqrt{(Q^2-4S)}}{2} \\
%Y&=&\frac{-H \pm \sqrt{H^2-4I}}{2}+e_{3y}+\rho_3 \sin(\theta)-e_{4y} \\
%Z&=& e_{3z}+ \rho_3 \cos(\theta)
%\end{eqnarray}
%Inverse kinematic equations can be obtained in the form
%\begin{eqnarray}
%\rho_1\!\!&=& \! X-e_{1x}-\frac{L3}{2}-\sqrt{L^2-(Y-e_{3y}-\tan(\theta)(Z-e_{3z})+e_{4y}-e_{1y})^2} \nonumber\\ 
%\rho_2\!\!&=& \! L_2-X-\frac{L3}{2}-e_{2x}-\sqrt{L^2-(Y-e_{3y}-\tan( \theta)(Z-e_{3z})+e_{4y}-e_{2y})^2 }\nonumber\\
%\rho_3\!\!&=& \! \frac{Z-e_{3z}}{\cos(\theta)}
%\end{eqnarray}
%%%%%%%%%%%%%%%%%%%%%%%%%%%%%%%%%%%%%%
\CCLsection{Workspace analysis}
%%%%%%%%%%%%%%%%%%%%%%%%%%%%%%%%%%%%%%
The workspace of a robot can be defined as the set of points that can be reached by its end‐effector (\cite{khalil2000modeling}). This is the space where the robot can operate and carry out tasks. In this work, the workspace is evaluated according to the reference point $V$ of the $\alpha$ mechanism. It is important to know the positions $V$ can reach to avoid issues such as singularity configurations, interference, collisions. Workspace is assessed as an intersection of two sets of points\footnote{See the movie: https://youtu.be/Y0uyeXuP5Eg}. %In order to design the workspace, two sets will be considered, first one belonging to a mechanism constituted by the links 1, 2, 3, and 4, and by the one formed by links 1, 4, 5 and 6, as shown in Figures XX and XX respectively. 
Following distances have to be defined:
\begin{itemize}
\item	$e_{5x}$: is the distance between the y-axis and the a joint, when the left actuator is at the end of its stroke;
\item	$e_{6x}$: is the distance between the y-axis and the e joint, when the right actuator is at the end of its stroke;
\item	$e_{7x}$: is the distance between the y-axis and the e joint when the right actuator is at the end of his stroke.
\end{itemize}

\begin{figure}[htbp]
        \centering
        \includegraphics[width = 0.7\textwidth]{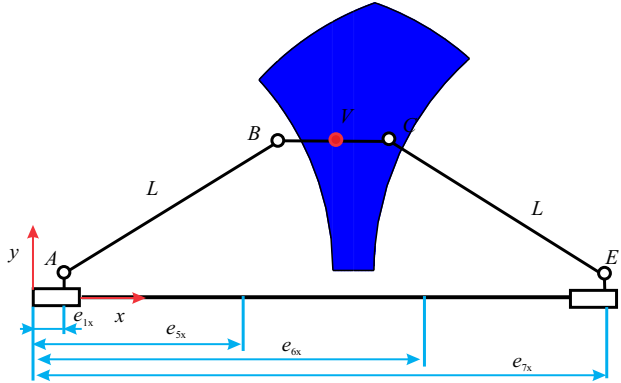}
        \caption{Workspace of the $\alpha$ mechanism.}
	    \label{fig:workspace_mechanism}
    \end{figure}

The complete workspace is displayed in Figure~\ref{fig:workspace_mechanism}. The workspace boundaries can be described analytically using five equations that can be used to verify whether the end-effector can reach a specified position or not.

\begin{figure}[htbp]
        \centering
        \includegraphics[width = 0.5\textwidth]{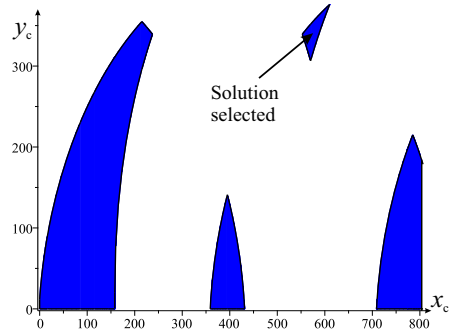}
        \caption{Possible placements of the regular workspace}
	    \label{fig:opt_solution}
    \end{figure}

\begin{figure}[htbp]
        \centering
        \includegraphics[width = 0.6\textwidth]{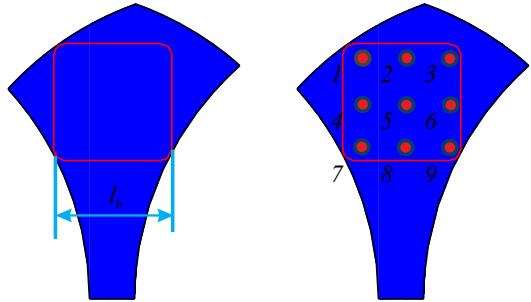}
        \caption{On the left, the regular square inside the regular workspace, on the right, the nine test points for the stiffness analysis with $x_c=570$ and $y_c=335$.}
	    \label{fig:regular_workspace_alpha_mechanism}
    \end{figure}

\begin{eqnarray}
\sqrt{(x-\frac{L_3}{2}-e_{1x})^2+(y-e_{1y})^2} \geq L, \quad 
\sqrt{(x+\frac{L_3}{2}-e_{7x})^2+(y-e_{2y})^2} \geq L \nonumber \\
\sqrt{(x-\frac{L_3}{2}-e_{5x})^2+(y-e_{1y})^2} \leq L, \quad
\sqrt{(x+\frac{L_3}{2}-e_{6x})^2+(y-e_{2y})^2} \leq L\nonumber 
\end{eqnarray}
To analyse the position of the end-effector, a regular workspace shape is considered. In our case, we evaluate the largest square that gives an inner tangent to the boundaries of the workspace (\cite{chablat2004interval}). The side length of the square is given by $l_b$ and, for algebraic computation, a Lame curve is used,

\begin{equation}
  (x-x_c)^n + (y-y_c)^n= l_b^n
\end{equation}
with $n$ a positive even integer and ($x_c,y_c$) are the center of the workspace. To limit the computation cost, $n$ is set at 6 as a first approximation and then to 12 to better approximate the square shape. 
To find the position of the regular workspace, the methodology defined by \cite{chablat2012solution} is used. The optimisation problem is written as the intersection of the robot workspace boundary with the boundary of the desired regular workspace whose placement depends on ($x_c, y_c$) . If the set of solutions obtained is non-zero, then we have found a candidate solution for the robot realization. This calculation was carried out using the Siropa library, whose functions are presented in (\cite{jha2018workspace}). To simplify the design, only commercial components from (\cite{makeblock}) have been used, without the need for machining. Several bar sizes were tested. This demonstrates that this robot can easily be built.  The main sizes are $L= 532$mm, $L_2=1300$mm, and $L_3=160$mm. For this set of design parameters, a set of possible placements for the regular workspace was found by using Cylindrical Algebraic Decomposition \cite{moroz2010cusp}. Figure~\ref{fig:opt_solution} presents several regions where there is no intersection between the regular workspace and the workspace of the robot. However, only one region is included in the workspace of the robot. 
If the robot can reach the desired workspace, we must be checked that the stiffness and workspace properties are under the specifications. The stiffness of the mechanism is therefore analysed in nine different configurations, which correspond to the precision points from 1 to 9, as shown in Figure \ref{fig:regular_workspace_alpha_mechanism}.
    
%%%%%%%%%%%%%%%%%%%%%%%%%%
\CCLsection{Stiffness analysis}
%%%%%%%%%%%%%%%%%%%%%%%%%%
To evaluate the stiffness of the robot to meet the specifications, simulations in CATIA were performed. The robot's stiffness is analysed in its working space. Two types of simulations have been taken into account: (i) when $\rho_1$ and $\rho_2$ move, i.e. the $\alpha$ mechanism, and (ii) when $\rho_3$ moves, i.e. the $\beta$ mechanism. The materials used are mainly aluminium, except for the shafts, the bearings, and slide guides, which are made of steel.  In all simulations, the self-weight of the structure is taken into account. 
As ball bearings are used, there is little backlash in the robot. The bending of the end-effector is due to the bending of the arms. The finite element analysis gives us the following results for case (i), Table~\ref{fig:ma_fig1} and case (ii), Table~\ref{fig:ma_fig2}. However, as we are using an external sensor-based controller, an embedded  camera, the most important issue is to have a constant deformation.

\begin{figure}[htbp]
        \centering
        \includegraphics[width = 0.85\textwidth]{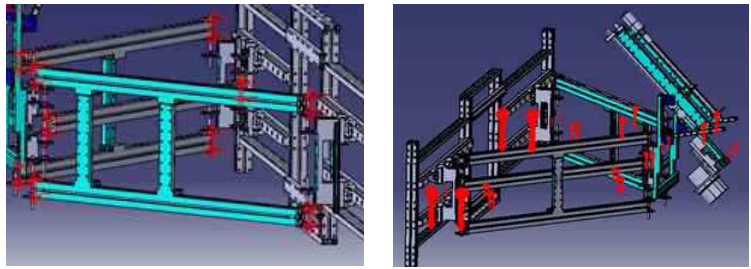}
        \caption{Boundaries conditions and gravity load in the CAD modelling}
	    \label{fig:stiffness}
\end{figure}

\begin{table}[t]
  \begin{minipage}[b]{0.50\linewidth}
  \center
\begin{tabular}{| c | c |}
\hline
Position	& Max displacement  [mm]\\  \hline
1	& 0.995 \\
2	& 0.993 \\
3	& 0.991 \\
4	& 0.990 \\
5	& 0.996 \\
6	& 0.989 \\
7	& 1.030 \\
8	& 1.010 \\
9	& 1.030 \\ \hline
\end{tabular}
  \caption{Maximum displacements throughout the workspace}
  \label{fig:ma_fig1}
  \end{minipage} 
  \begin{minipage}[b]{0.4\linewidth}
  \center
\begin{tabular}{| c | c |}
\hline
$\rho_3$ & Max displacement \\ 
$[mm]$ & [mm]\\ \hline
50	& 1.080 \\
75	& 0.996 \\
100	& 0.936 \\
125	& 0.907 \\
150	& 0.958 \\ \hline
\end{tabular}
  \caption{Maximum displacements for the $\beta$ mechanism}
  \label{fig:ma_fig2}
\end{minipage}
 \end{table}
%%%%%%%%%%%%%%%%%%%%%%%%%%%%%%%%%%%%%%
\CCLsection{Hardware and software}
%%%%%%%%%%%%%%%%%%%%%%%%%%%%%%%%%%%%%%
The control is realized using a Rasberry Pi 4 board, which allows the use of a touch screen, a camera to detect the position of the socket, and a serial link to an Arduino board (Microcontroller Me Auriga MakeBlock). Two stepper motors are used (42BYG Geared Stepper Motor) with two belts and four pulleys for the $\alpha$ mechanism, and a DC motor and a ball screw for the $\beta$ mechanism (Makeblock DC25 with sensor, 86 rpm). MakeBlock modules are used for the input-output connection with the Arduino board.
A Python program manages a user interface, the camera with OpenCV functions, and the communication with the Arduino board. To locate the plug on the car, we use a QR code stuck next to the plug. 
When the robot moves, the QR code seen by the camera is used to adjust the trajectory before starting the insertion of the plug\footnote{See the movie: https://youtu.be/P5wCgRqSyDQ}.

%Raspberry Pi 4 Model B
%MakeBlock Me Stepper Motor Driver
%MakeBlock Me RJ25 Adapter
%42BYG Geared Stepper Motor ratio  5.18:1, Phase Current: 2.8A
%Makeblock DC25" (86 rpm)
%%%%%%%%%%%%%%%%%%%%%%%%%%%%%%%%%%%%%%
\CCLsection{Conclusions}
%%%%%%%%%%%%%%%%%%%%%%%%%%%%%%%%%%%%%%
In this article, we presented a robot that allows the insertion of a plug into an electric car for charging. This robot is designed for the Zoe car but could be adapted to other vehicles having a front charging system, like the Nissan Leaf. To reduce manufacturing costs, off-the-shelf components were used in combination with hybrid kinematics with a simple kinematic model. A prototype was realized as a proof of concept associated with the patent application. A GitHub repository will soon be created to make the CAD and source code of the control available. 
%%%%%%%%%%%%%%%%%%%%%%%%%%%%%%%%%%%%%%
\CCLsection*{Acknowledgements}
%%%%%%%%%%%%%%%%%%%%%%%%%%%%%%%%%%%%%%
This work was supported by the project Chair between Renault and Centrale Nantes about performances of electric vehicles propulsion.
\bibliographystyle{plainnat}
\bibliography{Contribs}

\begin{thebibliography}{19}
\providecommand{\natexlab}[1]{#1}
\providecommand{\url}[1]{\texttt{#1}}
\expandafter\ifx\csname urlstyle\endcsname\relax
  \providecommand{\doi}[1]{doi: #1}\else
  \providecommand{\doi}{doi: \begingroup \urlstyle{rm}\Url}\fi

\bibitem[Chablat et~al.(2004)Chablat, Wenger, Majou, and
  Merlet]{chablat2004interval}
Damien Chablat, Ph~Wenger, F{\'e}lix Majou, and J-P Merlet.
\newblock An interval analysis based study for the design and the comparison of
  three-degrees-of-freedom parallel kinematic machines.
\newblock \emph{International Journal of Robotics Research}, 23\penalty0
  (6):\penalty0 615--624, 2004.

\bibitem[Chablat et~al.(2012)Chablat, Moroz, Arakelian, Briot, and
  Wenger]{chablat2012solution}
Damien Chablat, Guillaume Moroz, Vigen Arakelian, S{\'e}bastien Briot, and
  Philippe Wenger.
\newblock Solution regions in the parameter space of a 3-rrr decoupled robot
  for a prescribed workspace.
\newblock In \emph{Latest Advances in Robot Kinematics}, pages 357--364.
  Springer, 2012.

\bibitem[Figliolini et~al.(2012)Figliolini, Rea, and
  Angeles]{FiglioliniAsme2012}
Giorgio Figliolini, Pierluigi Rea, and Jorge Angeles.
\newblock The synthesis of the axodes of spatial four-bar linkages.
\newblock In \emph{International Design Engineering Technical Conferences and
  Computers and Information in Engineering Conference}, volume 45035, pages
  1597--1605. American Society of Mechanical Engineers, 2012.

\bibitem[Figliolini et~al.(2016)Figliolini, Rea, and
  Angeles]{FiglioliniRCC2016}
Giorgio Figliolini, Pierluigi Rea, and J.~Angeles.
\newblock The synthesis of the axodes of rccc linkages.
\newblock \emph{Journal of Mechanisms and Robotics}, 8\penalty0 (2):\penalty0
  021011, 2016.

\bibitem[Gao et~al.(2016)Gao, Mckay, Reiland, Foucault, Lacasse, Laliberte,
  Mayer-St-Onge, Lecours, Gosselin, Milburn, et~al.]{gao2016robotically}
Dalong Gao, Neil~David Mckay, Matthew~J Reiland, Simon Foucault, Marc-Antoine
  Lacasse, Thierry Laliberte, Boris Mayer-St-Onge, Alexandre Lecours, Clement
  Gosselin, David~E Milburn, et~al.
\newblock Robotically operated vehicle charging station, February~23 2016.
\newblock US Patent 9,266,440.

\bibitem[Jha et~al.(2018)Jha, Chablat, Baron, Rouillier, and
  Moroz]{jha2018workspace}
Ranjan Jha, Damien Chablat, Luc Baron, Fabrice Rouillier, and Guillaume Moroz.
\newblock Workspace, joint space and singularities of a family of delta-like
  robot.
\newblock \emph{Mechanism and Machine Theory}, 127:\penalty0 73--95, 2018.

\bibitem[Kamga(2017)]{Kamga:2017}
Gabriel~Simo Kamga.
\newblock Robotisation de la recharge à domicile d’une {R}enault {Z}o\'e.
\newblock Master's thesis, Ecole Centrale de Nantes, 2017.

\bibitem[Ketfi-C. et~al.(2018)Ketfi-C., Chablat, Wenger, and
  Ghanes]{KetfiCherif2018}
Ahmed Ketfi-C., Damien Chablat, Ph. Wenger, and M.~Ghanes.
\newblock Dispositif automatise de charge d'un véhicule électrique, 03 2018.
\newblock FR3078660.

\bibitem[Khalil and Dombre(2000)]{khalil2000modeling}
Wisama Khalil and Etienne Dombre.
\newblock Modeling, identification and control of robots, kp science, ed, 2000.

\bibitem[Majou et~al.(2002)Majou, Wenger, and Chablat]{majou2002design}
F{\'e}lix Majou, Philippe Wenger, and Damien Chablat.
\newblock Design of a 3 axis parallel machine tool for high speed machining:
  The orthoglide.
\newblock In \emph{IDMME, 4{\`e}me conf{\'e}rence internationale sur la
  conception et la fabrication int{\'e}gr{\'e}es en m{\'e}canique}, pages
  1--10. AIP-Prim{\'e}ca, 2002.

\bibitem[Makeblock(2022)]{makeblock}
Makeblock.
\newblock Makeblock®{US} official store, robot kits, stem toys for kids \&
  makers.
\newblock \url{http://https://www.makeblock.com/}, 2022.
\newblock Acc.: 21-02-16.

\bibitem[Miseikis et~al.(2017)Miseikis, Ruther, Walzel, Hirz, and
  Brunner]{miseikis20173d}
Justinas Miseikis, Matthias Ruther, Bernhard Walzel, Mario Hirz, and Helmut
  Brunner.
\newblock 3d vision guided robotic charging station for electric and plug-in
  hybrid vehicles.
\newblock \emph{arXiv preprint arXiv:1703.05381}, 2017.

\bibitem[Moroz et~al.(2010)Moroz, Chablat, Wenger, and Rouiller]{moroz2010cusp}
Guillaume Moroz, Damien Chablat, Philippe Wenger, and Fabrice Rouiller.
\newblock Cusp points in the parameter space of rpr-2prr parallel manipulators.
\newblock In \emph{New Trends in Mechanism Science}, pages 29--37. Springer,
  2010.

\bibitem[Ottaviano and Rea(2013)]{OttavianoRobotica2013}
Erika Ottaviano and Pierluigi Rea.
\newblock Design and operation of a 2-dof leg-wheel hybrid robot.
\newblock \emph{Robotica}, 31\penalty0 (8):\penalty0 1319--1325, 2013.

\bibitem[Rea et~al.(2013)Rea, Ottaviano, Conte, D’Aguanno, and
  De~Carolis]{Rea-IR2013}
Pierluigi Rea, Erika Ottaviano, Marco Conte, Angelino D’Aguanno, and Daniele
  De~Carolis.
\newblock The design of a novel tilt seat for inversion therapy.
\newblock \emph{International Journal of Imaging and Robotics}, 11\penalty0
  (3):\penalty0 1--10, 2013.

\bibitem[Walzel et~al.(2016)Walzel, Sturm, Fabian, and
  Hirz]{walzel2016automated}
Bernhard Walzel, Christopher Sturm, J{\"u}rgen Fabian, and Mario Hirz.
\newblock Automated robot-based charging system for electric vehicles.
\newblock In \emph{16. Internationales Stuttgarter Symposium}, pages 937--949.
  Springer, 2016.

\bibitem[Wenger et~al.(2001)Wenger, Gosselin, and
  Chablat]{wenger2001comparative}
Philippe Wenger, Cl{\'e}ment Gosselin, and Damien Chablat.
\newblock A comparative study of parallel kinematic architectures for machining
  applications.
\newblock \emph{Electronic Journal of Computational Kinematics}, 1\penalty0
  (1):\penalty0 23, 2001.

\bibitem[Yuan et~al.(2020)Yuan, Wu, and Zhou]{yuan2020concept}
Han Yuan, Qiong Wu, and Lili Zhou.
\newblock Concept design and load capacity analysis of a novel serial-parallel
  robot for the automatic charging of electric vehicles.
\newblock \emph{Electronics}, 9\penalty0 (6):\penalty0 956, 2020.

\bibitem[Zlatanov et~al.(1993)Zlatanov, Fenton, and
  Benhabib]{zlatanov1993singularity}
D~Zlatanov, RG~Fenton, and B~Benhabib.
\newblock Singularity analysis of mechanisms and robots via a velocity-equation
  model of the instantaneous kinematics.
\newblock In \emph{IEEE International Conference on Robotics and Automation},
  volume 1(2), pages 986--986. IEEE, 1993.

\end{thebibliography}

\end{document}